%% file: 0_Camera_Ready.tex
\documentclass[letterpaper]{article} 
\usepackage{aaai24}  
\usepackage{times}  
\usepackage{helvet}  
\usepackage{courier}  
\usepackage[hyphens]{url}  
\usepackage{graphicx} 
\usepackage{natbib}  
\usepackage{caption} 
\usepackage{algorithm}
\usepackage{amsmath}
\usepackage{algpseudocode}
\usepackage{listings}
\usepackage[dvipsnames]{xcolor}
\usepackage{makecell}
\usepackage{multirow}
\usepackage{subcaption}
\usepackage{url}
\usepackage{times}
\usepackage{helvet}
\usepackage{tabularx}  
\usepackage{array}
\usepackage{booktabs}

\usepackage{newfloat}
\usepackage{listings}
\DeclareCaptionStyle{ruled}{labelfont=normalfont,labelsep=colon,strut=off} 
\floatstyle{ruled}
\newfloat{listing}{tb}{lst}{}
\floatname{listing}{Listing}
\title{GameTileNet: A Semantic Dataset for Low-Resolution Game Art in Procedural Content Generation}
\author {
    Yi-Chun Chen\textsuperscript{\rm 1},
    Arnav Jhala\textsuperscript{\rm 2}
}
\affiliations {
    \textsuperscript{\rm 1}Yale University\\
    \textsuperscript{\rm 2}North Carolina State University\\
    ychen74@alumni.ncsu.edu, yi-chun.chen@yale.edu, ahjhala@ncsu.edu
}

\usepackage{bibentry}

\begin{document}

\maketitle

\begin{abstract}
GameTileNet is a dataset designed to provide semantic labels for low-resolution digital game art, advancing procedural content generation (PCG) and related AI research as a vision-language alignment task.
Large Language Models (LLMs) and image-generative AI models have enabled indie developers to create visual assets, such as sprites, for game interactions. However, generating visuals that align with game narratives remains challenging due to inconsistent AI outputs, requiring manual adjustments by human artists. 
The diversity of visual representations in automatically generated game content is also limited because of the imbalance in distributions across styles for training data. GameTileNet addresses this by collecting artist-created game tiles from OpenGameArt.org under Creative Commons licenses and providing semantic annotations to support narrative-driven content generation. The dataset introduces a pipeline for object detection in low-resolution tile-based game art (e.g., 32x32 pixels) and annotates semantics, connectivity, and object classifications. GameTileNet is a valuable resource for improving PCG methods, supporting narrative-rich game content, and establishing a baseline for object detection in low-resolution, non-photorealistic images.
\end{abstract}

\section{Introduction}
\input{1_Introduction}

\section{Related Work}
\input{2_RelatedWork}

\section{Dataset Overview and Annotation Schema}
\input{3_DatasetDescription}

\section{Dataset Statistics}
\input{4_Data_Statistics}

\section{Automatic Annotation Pipeline}

\input{5_Methodology}

\section{Narrative Scene Generation}
\input{6_Experiments_Results}
\section{Discussion and Conclusion}
\input{8_Conclusion}

\bibliography{aaai24}

\end{document}

%% file: 1_Introduction.tex
While Large Language Models (LLMs) hold great potential for generating diverse game narratives by enriching objects and scenes, translating these narrative elements into visual representations remains a challenge. Combining LLMs with visual models like diffusion models has produced appealing results, but aligning narrative and visual content in games often requires significant manual modifications and artistic judgment.
\cite{domsch2013storyplaying,shapiro2012realism,ang2006rules,koenitz2024narrative,williams2024tickrai}



Effective visual representations require integrating game materials with the narrative context to create immersive experiences for players. Therefore, visual elements like sprites and environments are essential for engagement and interaction.
The game world requires a variety of visual assets to align with narrative and player choices. 
\cite{bulatovic2024narrative,tyndale2016keys}


One potential solution is the semantic mapping between narrative elements and game materials. This approach ensures that visual assets align with narrative themes, allowing for the dynamic generation of visuals that reflect narrative chages. Semantic mapping helps maintain synergy between the game's visual and narrative aspects by streamlining the creation of procedurally generated content.



In this paper, we introduce GameTileNet, a dataset of game materials with semantic labels designed to facilitate such semantic mapping. Focused on top-down, tile-based pixel art, GameTileNet sources materials from OpenGameArt.org under Creative Commons licenses, providing a diverse range of assets. 



Pixel art tiles, typically low-resolution (e.g., 8x8, 16x16, or 32x32 pixels), use minimal memory and processing power, making them ideal for retro aesthetics. They rely on restricted color palettes to maintain visual consistency and their grid-based structure allows for modular game design, creating varied environments from a small set of assets.



The limited resolution and simplified details of pixel art make it difficult for computer vision models to distinguish subtle variations, classify objects, and accurately segment scenes. Furthermore, the niche nature of pixel art games means there is often a need for labeled data, hindering model training.



This study explores a pipeline for labeling tasks within this low-resolution dataset, focusing on pixel art game tiles. We demonstrate the use of GameTileNet through procedural content generation, using scene descriptions to create levels dynamically. The table \ref{teaser} showcases one of the results. These examples show how semantic mapping can align visual representations with generated narratives

\begin{table*}[!ht]
\centering
\renewcommand{\arraystretch}{1.1}
\setlength{\tabcolsep}{3pt}
\begin{tabular}{|
  m{3cm}|
  m{3cm}|
  >{\centering\arraybackslash}m{11cm}|}
\hline
\textbf{Place: Forest} & \textbf{Found Tiles} & \textbf{Showcase} \\
\hline
\begin{minipage}[c]{\linewidth}
\centering
\textcolor{red}{House} below \textcolor{green}{Tree}\\
\textcolor{green}{Tree} to the right of \textcolor{brown}{Barrel}\\
\textcolor{yellow}{Flower} above \textcolor{green}{Tree}\\
\textcolor{orange}{Tree stump} to the left of \textcolor{green}{Tree}
\end{minipage}
&
\begin{minipage}{\linewidth}
\centering
\setlength{\tabcolsep}{2pt}
\begin{tabular}{cc}
\includegraphics[height=1cm]{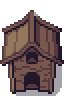} & 
\includegraphics[height=1cm]{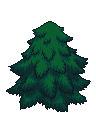} \\
House & Tree \\
\end{tabular}\\[2pt]
\begin{tabular}{cc}
\includegraphics[height=1cm]{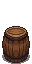} & 
\includegraphics[height=1cm]{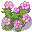} \\
Barrel & Flower \\
\end{tabular}\\[2pt]
\includegraphics[height=1cm]{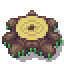}\\
Tree stump
\end{minipage}
&
\begin{minipage}[c]{0.48\linewidth}
\centering
\includegraphics[width=\linewidth]{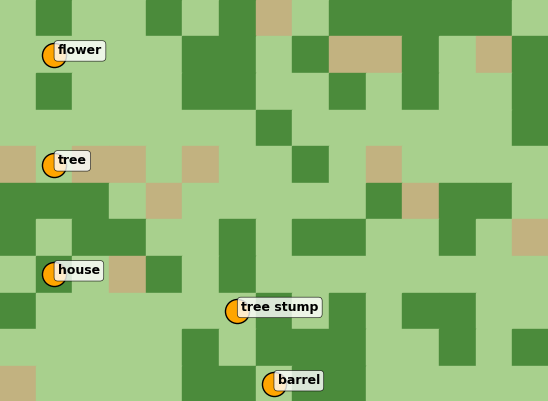}\\
\small Labeled layout
\end{minipage}
\hfill
\begin{minipage}[c]{0.48\linewidth}
\centering
\includegraphics[width=\linewidth]{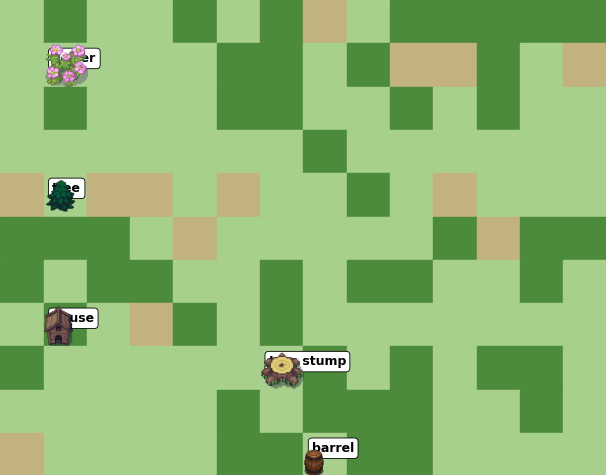}\\
\small Mapped tiles
\end{minipage}
\\
\hline
\end{tabular}
\caption{Create a game scene based on scene descriptions by aligning semantic labels in GameTileNet with narrative spatial relations. Left: labeled object layout. Right: mapped game tiles.}
\label{teaser}
\end{table*}

\begin{figure}[ht]
    \centering
    \includegraphics[width=0.45\textwidth]{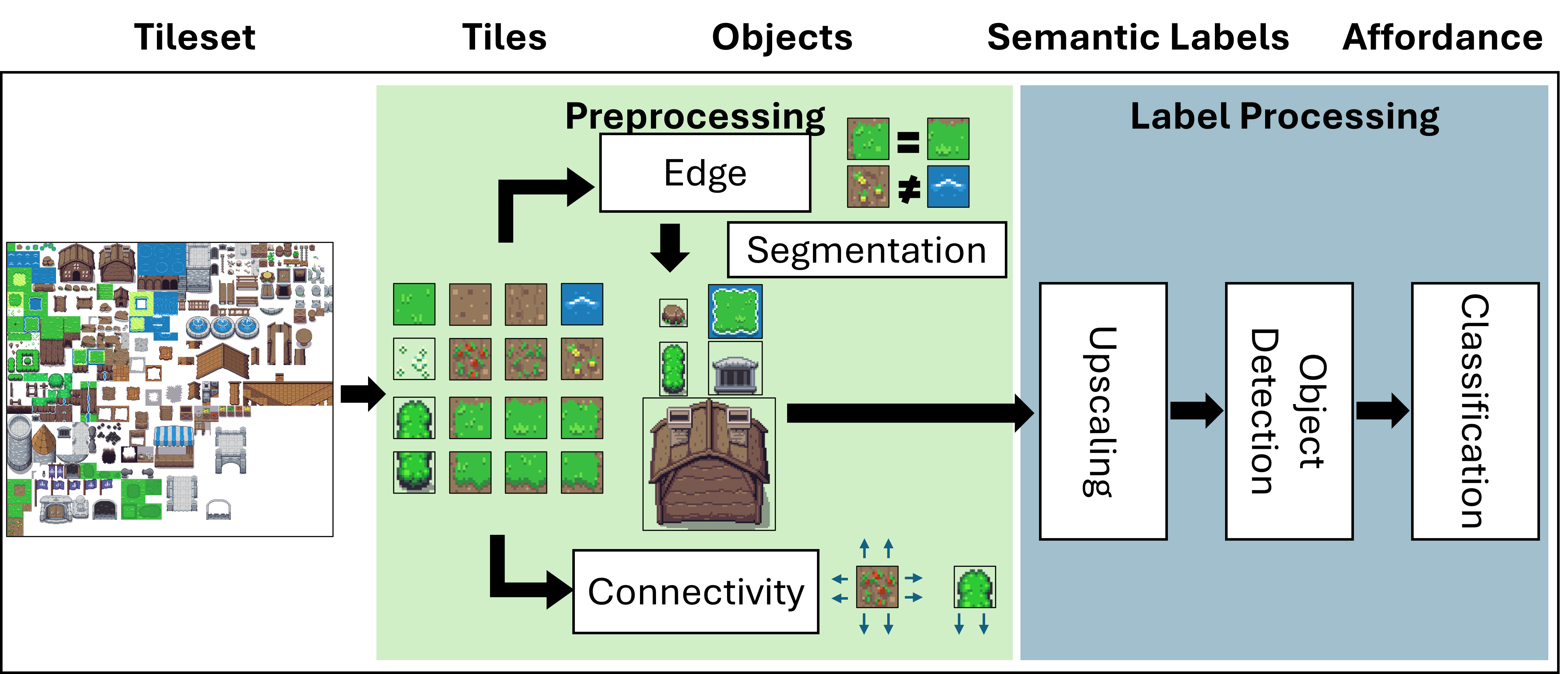}
    \caption{Object processing pipeline for pixel art game tiles. Each step supports accurate detection and labeling of low-resolution game elements.}
    \label{fig:auto_pipeline} 
\end{figure}

%% file: 2_RelatedWork.tex

\subsection{Tile Based Game Design and Semantics}
Tile-based game designs offer modular components that simplify game development by dividing environments into manageable units. This type of design supports efficient collision detection, level design, and resource optimization \cite{van2010tile,lafitte2009almost}. Beyond the structural role, tiles carry semantic meanings and affordances essential for guiding player interactions. Bentley and Osborn's Videogame Affordances Corpus illustrates how visual affordances signal potential actions, underscoring the need for tiles that reflect gameplay intent through semantic mapping \cite{bentley2019videogame}. Recent advancements, such as Jadhav and Guzdial's tile embeddings, provide affordance-rich representations that simplify embedding interactive elements in procedural generation \cite{jadhav2021tile}. At the same time, context-based segmentation systems enhance the alignment of game visuals with narrative cues using AI \cite{gabriel2023semantic}. Similar challenges of aligning visual assets with narrative structure have also been explored in comic generation systems using semantic layout and procedural asset control \cite{chen2024collaborative, chen2023customizable}. Tools like Tile Dreamer further streamline asset management, facilitating modular and dynamic level design \cite{karouzaki2007tile}. 

Grid-based procedural content generation (PCG) is a key for creating scalable, tile-based game environments. Foundational surveys by Togelius et al. and Hendrikx et al. explore search-based and modular PCG techniques that balance structured and random tile arrangements \cite{togelius2010search,hendrikx2013procedural}. As described by Yannakakis and Togelius, experience-driven PCG adapts content to player engagement \cite{yannakakis2011experience}. PCG via machine learning (PCGML) leverages neural networks to generate context-sensitive levels, with Guzdial et al. and Hu et al. showing how machine learning and large language models facilitate narrative-driven tile placement \cite{guzdial2022grid,hu2024game}. 
In addition, structured approaches to linking narrative events to visual layouts have been proposed in other domains such as comics, providing a conceptual foundation for applying narrative-based graphs to the game tile arrangement \cite{chen2025hierarchical}.

\subsection{The Challenge of Low-resolution Images}
Studies in low-resolution image recognition have shown that such low-detail characteristics impact the accuracy and usability of computer vision models. The study "80 Million Tiny Images" demonstrated that while a large dataset of low-resolution images can be beneficial for object recognition, limitations of the image details make distinguishing fine features difficult \cite{torralba200880}. After that, studies such as Wang et al. further highlight the performance drop on deep networks working with very low-resolution images (e.g., 16x16 pixels), where they proposed the use of domain adaptation and super-resolution methods to counterbalance resolution loss\cite{wang2016studying}. 
Semantic coherence and visual structure in low-resolution visual content have been addressed through hierarchical graph representations in visual narrative domains \cite{chen2025hierarchical}.



%% file: 3_DatasetDescription.tex

In this section, we describe our GameTileNet dataset and present descriptive statistics that characterize its expressive range.




\paragraph{Dataset Summary:}
GameTileNet includes 2,142 labeled game objects collected from top-down pixel art assets on OpenGameArt.org, a community-driven repository of open game materials. All assets are used under Creative Commons licenses, and attribution is preserved according to license terms. The dataset includes tiles from 67 tilesets and individual images created by 15 different artist groups.

As illustrated in Table~\ref{materialTypes}, game tile resources are available in two formats: full tilesets (which require segmentation) and individual object tiles. Each segmented tile is labeled with object names, semantic tags, connectivity, affordance types, and hierarchical metadata. Labels are provided by both annotators and original asset authors, supplemented by algorithmic methods.





\begin{table}[!ht]
\centering
\renewcommand{\arraystretch}{0.8} 

\begin{tabular}{|>{\centering\arraybackslash}m{3cm}|>{\centering\arraybackslash}m{3cm}|}
\hline
Tileset & Single Tile \\
\hline
\includegraphics[width=2cm]{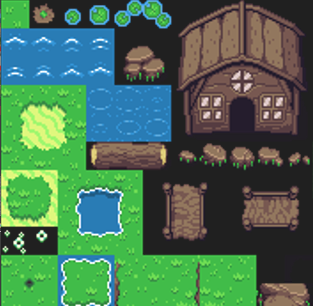} 
&
\includegraphics[width=2cm]{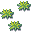} \\ 
\hline
\end{tabular}
\caption{Two types of game tile materials in OpenGameArt.org: a tileset and a single image.}
\label{materialTypes}
\end{table}


\paragraph{Creative Commons License:}
All images included in GameTileNet are used under one of three Creative Commons licenses: CC0, CC-BY (3.0 or 4.0), or CC-BY-SA (3.0 or 4.0).%
We ensure full compliance with all license terms, including attribution and share-alike requirements when applicable.



 

\subsection{Annotation Schema and Object Categories}

Our annotation schema focuses on describing tile content to support narrative mapping and scene composition. Labels originate from asset filenames, author-defined tags, and human annotations. After preprocessing, we normalized 315 author names and 86 annotator-provided tags into a unified set of 361 object names.

\subsection{Affordance Taxonomy and VGDL Alignment}

Our affordance labels are informed by the Video Game Description Language (VGDL)~\cite{schaul2013video}, which categorizes objects by their gameplay roles. We adapt this taxonomy through a visual-semantic lens suitable for 2D tile-based scenes. The five final categories are:

\begin{itemize}
  \item \textbf{Terrain:} Walkable or background surfaces such as grass or stone floors.
  \item \textbf{Environmental Object:} Static elements that provide context, including trees, fences, or architectural features.
  \item \textbf{Interactive Object:} Elements that players can trigger or move through, such as doors, levers, or switches.
  \item \textbf{Item/Collectible:} Objects that players can acquire or use, including potions, weapons, or scrolls.
  \item \textbf{Character/Creature:} Playable avatars, non-playable characters (NPCs), and enemies like goblins or shopkeepers.
\end{itemize}

\subsection{Annotation Schema: Tile Types and Segment Labels}
\label{define:usable}
Each segmented tile is labeled according to both its visual integrity and its semantic role. The annotation types are as follows:

\begin{itemize}
  \item \textbf{Tiles:} 32$\times$32 pixel units extracted from larger tilesets.
  \item \textbf{Complete:} Tiles containing entire objects, visually enclosed with no cropping at the edges.
  \item \textbf{Partial:} Tiles showing truncated or occluded objects that extend beyond the tile boundary.
  \item \textbf{Texture:} Repeating material surfaces, such as wood grain or patterned stone.
  \item \textbf{Complete\_texture:} Fully enclosed and tileable texture regions suitable for procedural repetition.
  \item \textbf{Partial\_texture:} Texture samples that are visually cropped or unsuitable for seamless repetition.
\end{itemize}

We define usable tiles as either complete visual entities or texture patches that are structurally appropriate for terrain composition.

\subsection{Core Annotation Tasks}

Building on the defined schema, we structure the annotation process around four core tasks:

\begin{itemize}
  \item \textbf{Adjacency (Edge Similarity):} Annotators judge if adjacent tile edges belong to the same object. Binary labels were collected for 200 randomly sampled tile pairs.
  \item \textbf{Tile Connectivity:} Annotators specify the directions (e.g., top-right, top-left) a tile can connect to others. This supports tile-based map assembly.
  \item \textbf{Object Detection:} Tiles are labeled with semantic object names.
  \item \textbf{Object Categorization:} Names are grouped into functional categories to reduce label noise and improve generalization.
\end{itemize}

While fixed vocabularies simplify labeling, subjective cases (e.g., partial crops) remain challenging. Interfaces used in these tasks are shown in Table~\ref{LabelingInterface}.

\begin{table}[!ht]

\centering
\begin{tabular}{|c|c|c|}
\hline
Object Labels & Connectivity & Similarity \\
\hline
\includegraphics[width=0.12\textwidth]{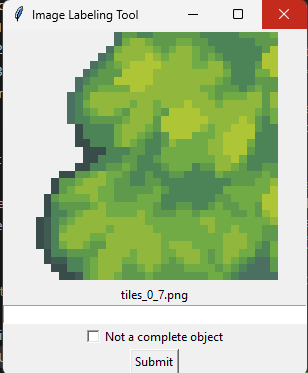} &
\includegraphics[width=0.12\textwidth]{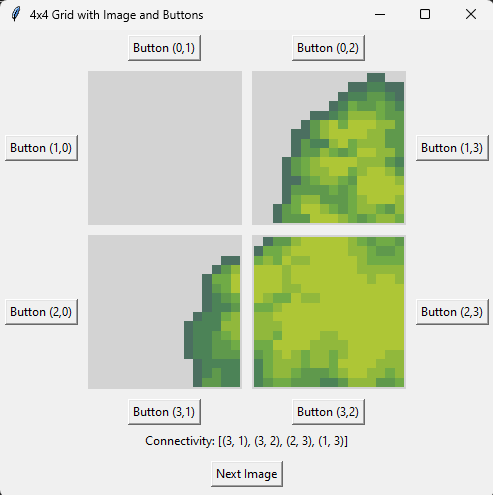} &
\includegraphics[height=0.14\textwidth]{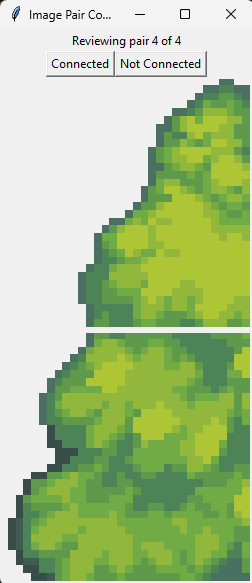} \\
\hline
\end{tabular}
\caption{Labeling interfaces for different tasks.}
\label{LabelingInterface}
\end{table}


\vspace{-0.5cm}
\paragraph{Consistency Assessment}
To evaluate annotation consistency, we performed a relabeling experiment. A subset of 50 tiles was randomly selected from the validation set and re-annotated by the same annotators after a two-week interval. 

\subsection{Author-Labeled Object Tiles}

We include 1,898 additional object tiles originally labeled by asset creators. These objects come with free-text names and categories reflecting fantasy game conventions. The distribution includes: \textbf{monsters} (1,012), \textbf{items} (732), \textbf{tiles} (91), \textbf{terrain} (33), \textbf{player characters} (29), and \textbf{beings} (10).

To ensure consistency, we group these into six normalized categories—\textit{creatures}, \textit{items}, \textit{player units}, \textit{tiles}, \textit{terrain}, and \textit{narrative agents}—and standardize their labels.

\subsection{Hierarchical Semantic Labeling Schema}

Each object in the dataset is annotated with a four-level semantic schema, inspired by layered scene annotation practices such as ADE20K~\cite{zhou2017scene,zhou2019semantic}:

\begin{itemize}
  \item \textbf{Detailed Name:} The specific identifier for an object, often derived from file names or author labels.
  \item \textbf{Group Label:} A normalized, canonical name representing object variants under a single term.
  \item \textbf{Supercategory:} A higher-level classification that groups related objects, such as terrain or character types.
  \item \textbf{Affordance Label:} The gameplay function of the object, informed by VGDL-style interaction roles.
\end{itemize}

This schema enables both fine-grained and abstract reasoning across visual game elements and supports affordance-aware detection and symbolic alignment tasks.

%% file: 4_Data_Statistics.tex



\begin{figure*}[t]
  \centering
  \begin{subfigure}[b]{\textwidth}
    \centering
    \includegraphics[width=\linewidth]{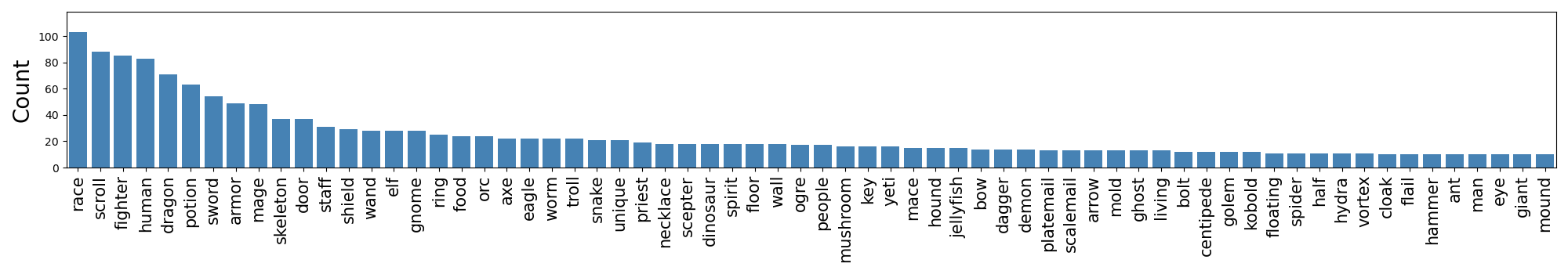}
    \caption{Object counts for each group label, sorted by frequency.}
    \label{fig:sub_group_counts}
  \end{subfigure}

  \vspace{1em} 

  \begin{subfigure}[b]{\textwidth}
    \centering
    \includegraphics[width=\linewidth]{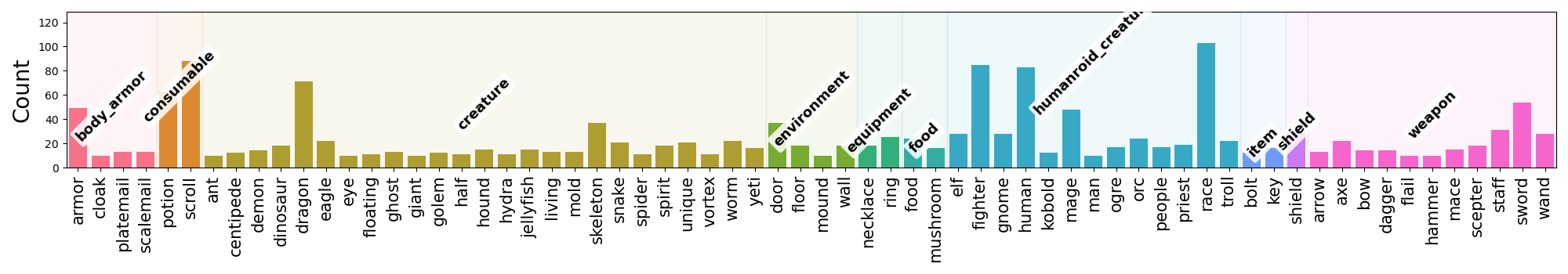}
    \caption{Group labels clustered by supercategories, with color-coded class regions.}
    \label{fig:sub_group_supercategory}
  \end{subfigure}

  \caption{Distribution and semantic structure of object categories in GameTileNet. (a) Group label frequencies, limited to labels with more than 10 occurrences. (b) Supercategories over the full set; colored regions show dataset-wide coverage.}
  \label{fig:group_distribution_combined}
\end{figure*}

Table \ref{tab:segmentation_summary} give the statistics of usable units from 10995 game tiles, defined by tile completeness. Table \ref{labeledStatistics} presents the statistics of annotated data, including the Adjacency (Edge Similarity) 
and Connectivity of game tiles, the object names of tile content, and the classes to which the objects belong. Additionally, subfigures \ref{fig:sub_group_counts} and \ref{fig:sub_group_supercategory} illustrate the frequency of objects appearing in GameTileNet. 
\begin{table*}[ht]

\centering
\renewcommand{\arraystretch}{1.1}
\begin{tabularx}{\linewidth}{|l|X|X|X|X|X|X|}
\hline
\textbf{Tileset} & \textbf{Total} & \textbf{Partial} & \textbf{Complete} & \textbf{P\_texture} & \textbf{C\_texture} & \textbf{Usable\_rate} \\
\hline
Segment\_algo\_recursive & 1818 & 1064 & 737 & 279 & 17 & 56.8\% \\
\hline
Segment\_model & 2612 & 1754 & 841 & 17 & 171 & 39.4\% \\
\hline
\end{tabularx}
\caption{Summary of tile segment counts and usable rate from two segmentation methods. The usable rate measures the proportion of extracted segments that result in complete, recognizable objects or textures (e.g., a full chest) rather than partial or cropped fragments.}
\label{tab:segmentation_summary}
\end{table*}


\begin{table*}[!ht]

    \centering
    \small
    \begin{tabular}{|c|c|c|c|c|c|c|}
        \hline
        & Adjacency & Connectivity & \# Objects Labeled & Object Groups & \# Supercategories & Affordance Labeled \\
        \hline
        Annotator & 200 & 100 & 244 & 86 & 26 & 2142 \\ \hline
        Author & X & X & 1898 & 315 & X & X \\ \hline
        Algo \& Models & 90.5\% & 43\% & X & X & X & See Table~\ref{tab:affordance_metrics} \\ \hline
    \end{tabular}
    \caption{
        Summary of annotation statistics from three sources: human annotators, original content authors, and automatic models. "Adjacency" and "Connectivity" refer to tile relationships. "Affordance Labeled" indicates the number of tiles annotated with affordance types or prediction accuracy (detailed in Table~\ref{tab:affordance_metrics}).
    }
    \label{labeledStatistics}
\end{table*}

\subsubsection{Affordance Label Distribution Analysis}

To understand the diversity and overlap of affordance types in our dataset, we analyzed the annotated labels from two perspectives: (1) treating each unique combination of affordances as a single class, and (2) counting each individual label occurrence independently.

\begin{figure*}[ht]
    \centering
    \begin{subfigure}[t]{0.45\textwidth}
        \centering
        \includegraphics[width=\textwidth]{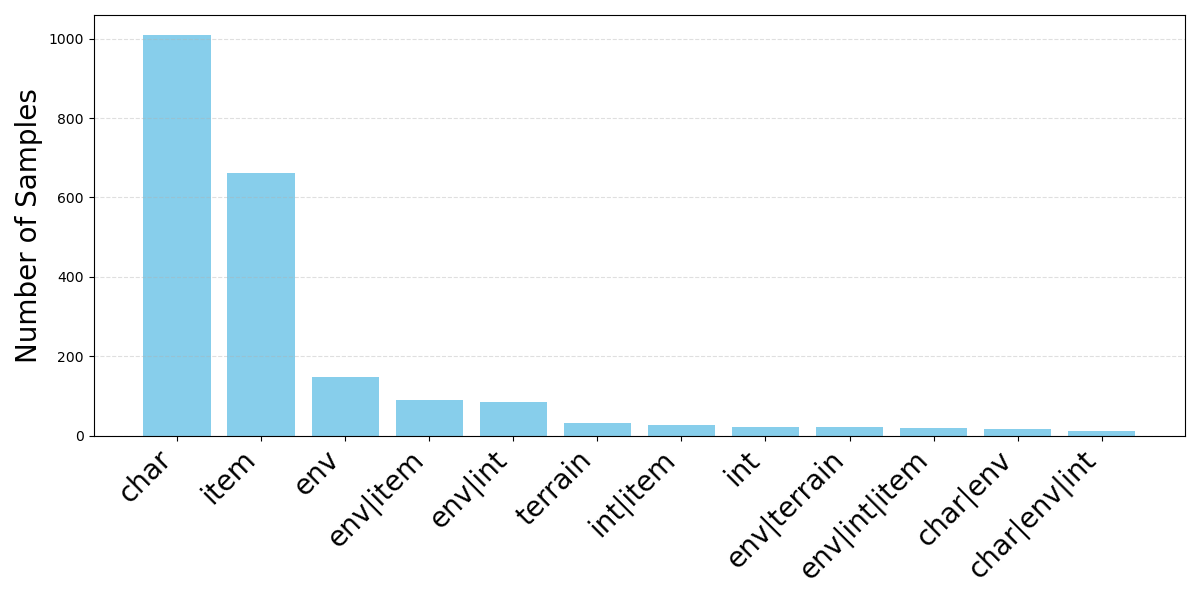}
        \caption{Combined affordance label distribution. Each bar is a unique label combination.}
        \label{fig:flaten_affordance_combined}
    \end{subfigure}
    \hfill
    \begin{subfigure}[t]{0.45\textwidth}
        \centering
        \includegraphics[width=\textwidth]{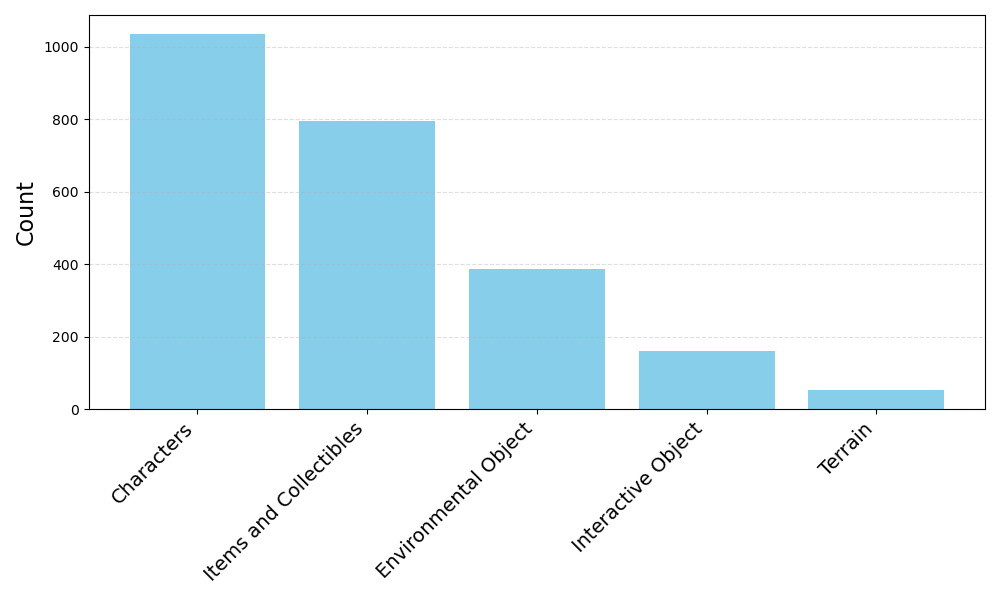}
        \caption{Separated label distribution. Labels are split and counted independently.}
        \label{fig:flaten_affordance_separate}
    \end{subfigure}
    \caption{
        Affordance label distributions across annotated object tiles. 
        \textbf{Left:} Unique label combinations show the diversity of functional roles. 
        \textbf{Right:} Individual label counts reveal frequent co-occurrence among affordances, except for \texttt{Characters}, which rarely overlaps.
    }
    \label{fig:affordance_label_distributions}
\end{figure*}

\paragraph{Combined Label View.}
Figure~\ref{fig:flaten_affordance_combined} (left) shows the distribution of combined affordance labels. The most frequent class is \texttt{Characters}, followed by \texttt{Items and Collectibles} and multi-label combinations such as \texttt{Environmental Object | Items and Collectibles}. This view highlights how often objects serve multiple gameplay functions, especially among background elements or items.

\paragraph{Separated Label View.}
Figure~\ref{fig:flaten_affordance_separate} (right) presents the distribution when labels are split and counted individually. Here, \texttt{Characters} remains dominant, but we also observe a high frequency of \texttt{Items and Collectibles}, \texttt{Environmental Object}, and \texttt{Interactive Object}. This reveals significant label co-occurrence, except for \texttt{Characters}, which tends to appear alone.

\paragraph{Affordance Overlap Insight.}
These two views together confirm that many game objects have overlapping functional roles—such as being both interactable and collectible—while \texttt{Characters} remain semantically distinct. This overlapping structure has important implications for downstream modeling.

\paragraph{Affordance Prediction as a Multi-Label Classification Task}

The affordance annotations in our dataset reflect a multi-label structure—many objects are associated with more than one gameplay function. 
To preserve this structure, we formulate affordance prediction as a \textbf{multi-label classification problem}, where each object can be assigned multiple affordance labels. This approach better reflects real-world usage and supports flexible downstream applications such as gameplay reasoning and adaptive object interaction.

We adopt a lightweight model using CLIP embeddings and a multi-label MLP classifier. The training pipeline supports multi-hot targets and uses binary cross-entropy loss to optimize prediction across all label dimensions. 

%% file: 5_Methodology.tex
We design a compact pipeline for object detection in low-resolution pixel art tiles (Figure~\ref{fig:auto_pipeline}). Each tileset is first split into 32$\times$32 units, which are grouped by visual similarity and summarized into a compact dictionary for efficient matching. Optional background padding is added to enhance context, and tiles are embedded into feature vectors using pretrained visual encoders (e.g., CLIP). To improve clarity, images are upscaled before being passed to object detection models, whose outputs are used to assign affordance labels.

To support reuse of GameTileNet in PCG tasks, we evaluate the quality and robustness of each pipeline stage: segmentation, connectivity, and affordance prediction, rather than comparing against a universal baseline (which does not exist for low-resolution tile art). This component-wise validation ensures the pipeline produces reliable labels that can generalize to unseen tiles.



\paragraph{Adjacency (Edge Similarity):}
To determine whether adjacent tiles form a connected pair, we measure structural similarity at their shared boundaries. Given two adjacent tiles, we extract the connecting boundary rows or columns, convert them to grayscale (removing color variation), and compute the Structural Similarity Index Measure (SSIM)~\cite{wang2004image}. SSIM evaluates luminance ($\mu$), contrast ($\sigma$), and structural correlation ($\sigma_{B_1 B_2}$), offering a perceptually motivated metric that outperforms raw pixel comparison. Its value ranges from $-1$ (structurally different) to $1$ (identical). The similarity between two boundary regions $B_1$ and $B_2$ is defined as:
\[
 SSIM(B_1, B_2) =
 \frac{(2\mu_{B_1}\mu_{B_2} + C_1)(2\sigma_{B_1 B_2} + C_2)}
      {(\mu_{B_1}^2 + \mu_{B_2}^2 + C_1)(\sigma_{B_1}^2 + \sigma_{B_2}^2 + C_2)}.
\]




\paragraph{Segmentation:}

For segmenting objects inside tile sets, we employ a recursive region-growing approach that detects connected segments based on adjacency and similarity. 
The process begins with an unvisited tile, from which the algorithm recursively explores all four neighboring tiles (right, left, down and up), forming a connected segment. A tile is considered part of the same segment if it meets two conditions: (1) it is not primarily transparent, ensuring that only meaningful tiles are included, and (2) it passes a similarity threshold check with its adjacent tile. Each segment is anchored by its first encountered tile, designated as the parent, which groups all connected tiles under it. 

\paragraph{Dataset and Training}
We curated a binary classification dataset with thousands of labeled tile segments. A lightweight ResNet18 model pretrained on ImageNet was fine-tuned on these examples. Tiles were resized to 64$\times$64, and the data was split into training, validation, and test sets (70\%\/15\%\/15\%). Training used cross-entropy loss, the Adam optimizer, and early stopping with a maximum of 15 epochs.

\paragraph{Performance}
We conducted three training runs to evaluate stability. The best model achieved 92.2\% test accuracy, with an average validation accuracy of 91.8\%. Accuracy typically plateaued by epoch 8--10, demonstrating efficient convergence and minimal overfitting.

\begin{figure}[!ht]
    \centering
    \includegraphics[width=0.25\textwidth]{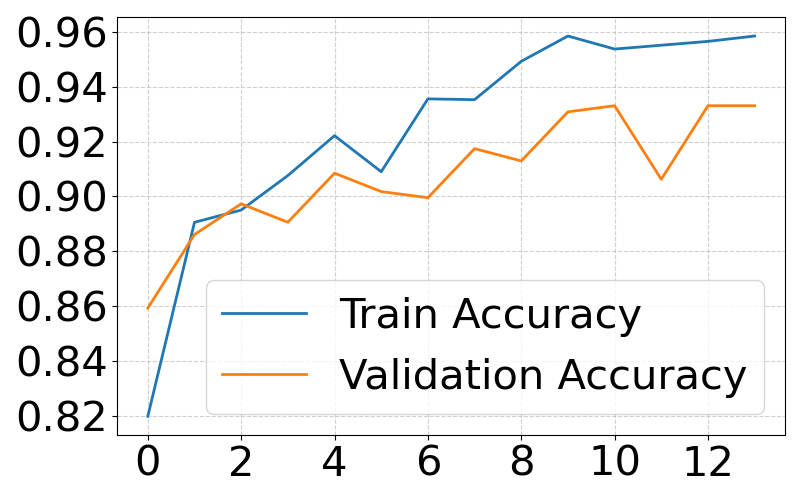}
    \caption{Training and validation accuracy across epochs for the completeness classification task (y-axis: accuracy, x-axis: epochs).}
    \label{fig:completeness_training_curve}
\end{figure}


\paragraph{Discussion}
The classifier reliably distinguishes complete from partial tiles using low-level visual features. This supports automated pre-filtering of segments for reuse, reducing manual annotation effort and improving scalability for applying our framework to new game assets.

\paragraph{Connectivity}
Tile-based games require seamless transitions between adjacent tiles to ensure visually coherent environments. We define tile connectivity in terms of directional adjacency, where each tile is divided into eight directional edge segments (e.g., top-left, right-down). We propose a hybrid rule-based algorithm that combines visual similarity and edge transparency to infer directional connectivity between tiles. For each tile and each edge direction, we first check whether the corresponding edge segment is non-transparent, based on the alpha channel of the image. If the edge is non-transparent, it is marked as a potential connection. If a neighboring tile exists in that direction, we extract the overlapping edge regions from both tiles and compute the Structural Similarity Index Measure (SSIM). If the SSIM score exceeds a predefined threshold (e.g., 0.6), the edge is considered a valid directional connection. This dual-filtering approach helps avoid both false connections from transparent edges and false negatives due to visual variation.

\paragraph{Comparison with Manual Annotations}

        
            
                
                    
        
To assess the quality of automatically generated connectivity labels, we compared the predicted edges against manually annotated ground truth using precision, recall, F1 score, and exact match rate. The best-performing variant achieved an average F1 score of 0.63 and an exact match rate of 43\%, indicating a moderate level of agreement with human annotations. 
The balanced precision and recall further support the trustworthiness of the predicted labels for downstream tasks. We adopt this version as our refined connectivity annotation baseline and use it in subsequent experiments.

\paragraph{Semantic Labels}
The dataset includes two types of semantic labels: (1) object names, and (2) categorical classes. To normalize game-related words into categories such as Characters, Items, Terrain, Non-human, and Attribute, annotators used GPT-4 for suggestions in ambiguous cases (e.g., when file names were unclear), then verified or corrected the results. Although not formally benchmarked, this process improved consistency and reduced label noise.

\subsubsection{Preprocessing}
In the preprocessing steps, we utilize the Natural Language Toolkit (NLTK) for initial cleaning \cite{bird2009natural,loper2002nltk}. The cleaning process includes three steps: first, checking for and removing duplicated words; second, filtering the words to retain only nouns; and third, identifying and eliminating plurals and transformed forms. {\em Environmental Objects:} Tree, Rock, Water, Grass, Pathway, Building. {\em Interactive Objects: } Door, Key, Chest, Switch, Lever. {\em Characters and NPCs:} Player, Enemy, NPC, Boss. {\em Items and Collectibles: } Potion, Coin, Gem, Weapon, Armor

\subsubsection{Tile Upscaling Methods for Visual Enhancement}

Low-resolution tiles often suffer from aliasing and loss of detail, which hinders accurate object detection. We evaluate five upscaling methods to enhance 32$\times$32 pixel game tiles prior to recognition:

\begin{itemize}
    \item \textbf{Bicubic Interpolation (PIL):} A baseline method using traditional image resizing via Python Imaging Library.
    \item \textbf{ESRGAN:} A generative adversarial network trained for perceptual super-resolution on natural images, preserving sharp edges and textures.
    \item \textbf{SwinIR:} A transformer-based method designed for image restoration tasks, which balances fidelity and structure preservation.
    \item \textbf{Stable Diffusion (Fidelity):} Image super-resolution using diffusion models conditioned on the original image for high-detail fidelity.
    \item \textbf{Stable Diffusion (Img2Img):} A creative variant of Stable Diffusion that allows semantic enhancement via denoising strength and prompt guidance.
\end{itemize}

Figure~\ref{table:upscaling_comparison} illustrates visual differences across methods on the same input tile.

\begin{table*}[h!]

\centering
\renewcommand{\arraystretch}{1.2}
\setlength{\tabcolsep}{4pt}
\begin{tabular}{|>{\centering\arraybackslash}m{2.5cm}|
                >{\centering\arraybackslash}m{2.5cm}|
                >{\centering\arraybackslash}m{2.5cm}|
                >{\centering\arraybackslash}m{2.5cm}|
                >{\centering\arraybackslash}m{2.5cm}|
                >{\centering\arraybackslash}m{2.5cm}|}
\hline
Original & Bicubic (PIL) & ESRGAN & SwinIR & Stable Diffusion (Fidelity) & Stable Diffusion (Img2Img) \\
\hline
\includegraphics[width=2.3cm]{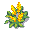} &
\includegraphics[width=2.3cm]{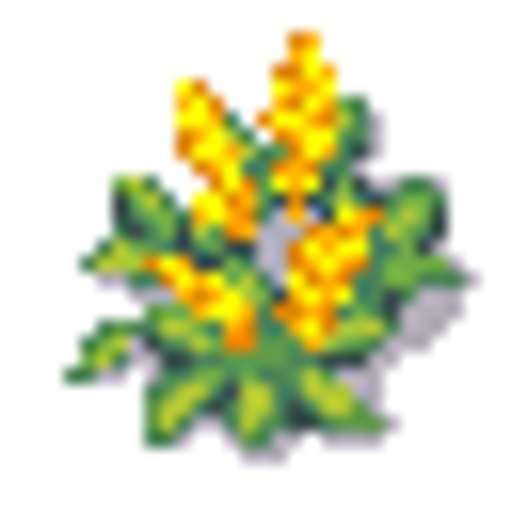} &
\includegraphics[width=2.3cm]{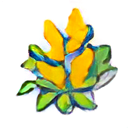} &
\includegraphics[width=2.3cm]{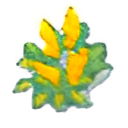} &
\includegraphics[width=2.3cm]{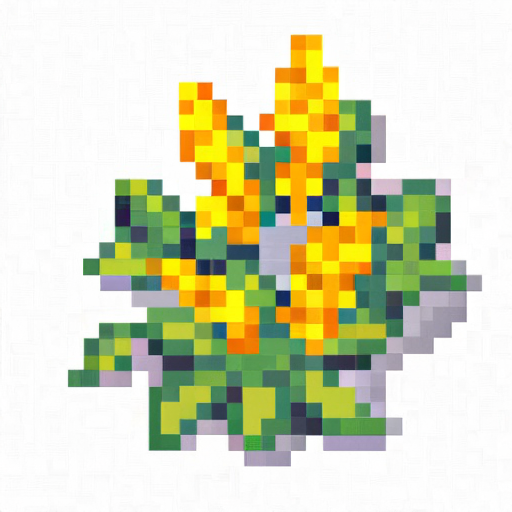} &
\includegraphics[width=2.3cm]{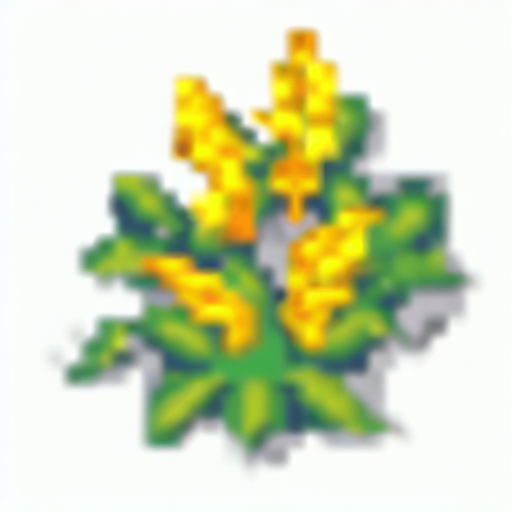} \\
\hline
\end{tabular}
\caption{Visual comparison of five upscaling techniques on a 32$\times$32 tile (Environmental Object: yellow flower on grass). Stable Diffusion is shown in both fidelity-preserving and generative (img2img) variants.}
\label{table:upscaling_comparison}
\end{table*}

\subsubsection{Detection and Semantic Matching}
We evaluate semantic understanding of game tiles using two approaches: a conventional object detector (YOLOv5s) and a vision-language captioning model (BLIP). Both are applied to original and upscaled images to assess the impact of enhancement on recognition.

YOLOv5s, pretrained on COCO, provides bounding-box detections with fixed real-world labels, which are poorly suited for fantasy-themed, low-resolution tiles. While upscaling improves detection quantity, domain mismatch prevents alignment with our semantic categories. In contrast, BLIP generates free-form captions in an open-vocabulary setting, enabling recognition of stylized or unseen objects. We compare these captions to annotated group labels, supercategories, and affordances to assess semantic alignment.

Overall, YOLO offers spatial grounding but struggles with vocabulary fit, while BLIP supports descriptive and domain-adaptive recognition without localization. BLIP’s outputs show promise for bootstrapping semantic labels, clustering assets, or supporting interactive authoring in creative workflows.

\subsubsection{Evaluating Caption–Label Similarity}

To assess whether vision-language captioning methods such as BLIP can help recognize the semantics of game tiles, we compare the generated captions with curated object labels. Each image is associated with three types of ground-truth labels: (1) \textbf{group labels}, (2) \textbf{supercategories}, and (3) \textbf{affordance labels}. These labels were manually annotated by authors or external annotators.

We evaluate the similarity between generated captions and ground-truth labels along three levels: \textbf{Direct Match:} We check whether any exact word in the caption matches a word from the group labels, supercategories, or affordances. \textbf{Synonym Match:} Using WordNet-based synonym sets, we expand each label to include common synonyms and check if any appear in the caption. \textbf{Semantic Similarity:} We encode both the caption and each label term using a Sentence-BERT encoder and compute cosine similarity scores. A match is considered positive if the similarity score exceeds a threshold (e.g., 0.3). 
    
    
These three matching levels help evaluate whether captions preserve or convey the intended semantic information of the game tile images, and which upscaling method supports this best. The final results are reported as counts of matches across all images per label type and match level.

\begin{table*}[ht]

\centering
\small
\setlength{\tabcolsep}{3pt}
\begin{tabular}{|p{2.2cm}|c|c|c|c|c|}
\hline
\textbf{Image Source} &
\textbf{Total Img.} &
\textbf{YOLO Det. (\#)} &
\textbf{Dir. Match (G/S/A)} &
\textbf{Syn. Match (G/S/A)} &
\textbf{Sem. Match (G/S/A)} \\
\hline
Original (32x32) &
1898 &
N/A &
104/3/-- &
143/4/24 &
207/177/122 \\
\hline
Bicubic &
1898 &
1058 &
166/8/-- &
197/9/56 &
335/338/105 \\
\hline
Real-ESRGAN &
1898 &
1138 &
170/2/-- &
204/22/311 &
432/383/374 \\
\hline
SwinIR &
1889 &
1306 &
118/15/-- &
149/20/181 &
276/300/214 \\
\hline
SD (fidelity) &
1898 &
35 &
51/2/-- &
68/2/2 &
96/181/18 \\
\hline
SD (img2img) &
1898 &
1171 &
146/9/-- &
192/10/45 &
278/333/102 \\
\hline
\end{tabular}
\caption{Combined caption–label match statistics and YOLO detection counts across upscaling methods. G = group label, S = supercategory, A = affordance label. Although YOLO detections increased after upscaling (e.g., 1306 detections for SwinIR), none matched the target labels, highlighting a vocabulary mismatch.}
\label{tab:caption_yolo_combined}
\end{table*}

\begin{figure*}[ht]
    \centering
    \includegraphics[width=0.75\textwidth]{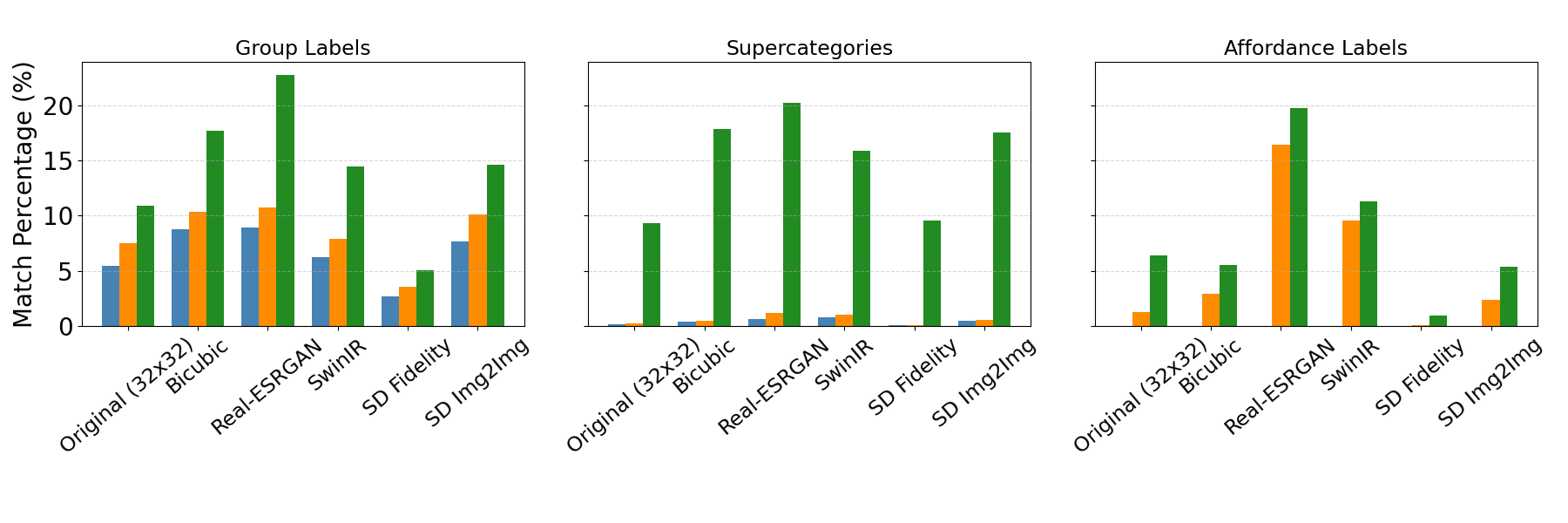}
    \caption{
    Percentage of caption–label matches across different upscaling methods. 
    Each bar shows the proportion of matches between BLIP-generated captions and author-annotated labels: 
    \textbf{Group Labels}, \textbf{Supercategories}, and \textbf{Affordance Labels}. 
    We measure three types of matches—
    \textcolor{blue}{\textbf{Direct Match}}, 
    \textcolor{orange}{\textbf{Synonym Match}}, and 
    \textcolor{green!50!black}{\textbf{Semantic Similarity}}—
    using lexical and embedding-based comparisons. 
    SwinIR and Real-ESRGAN upscaled images show notably better alignment with captions, especially in semantic terms.
    }
    \label{fig:caption_label_match}
\end{figure*}

\subsubsection{Evaluating Object Recognition and Semantic Label Alignment}

To assess how different upscaling methods affect the interpretability of game tiles, we evaluate two forms of automatic annotation: (1) object detection using YOLOv8 and (2) caption–label alignment using BLIP-generated image captions. Table~\ref{tab:caption_yolo_combined} summarizes the detection counts and match statistics, while Figure~\ref{fig:caption_label_match} visualizes the proportion of caption–label matches by match type and label type.

\paragraph{YOLO Detection and Its Limitations.}
Despite upscaling improving YOLO detection counts—for instance, SwinIR yields 1306 detections and Real-ESRGAN 1138 compared to none at 32×32 resolution—the matches against our annotated labels remain \textit{zero}. This reveals a fundamental limitation: YOLO models are trained on fixed, real-world categories (e.g., \textit{traffic light}, \textit{bicycle}), which rarely overlap with our fantasy-themed or stylized game objects. As a result, while YOLO detects more shapes after upscaling, these labels are largely irrelevant for domain-specific semantic understanding.

\paragraph{Caption-Based Semantic Matching.}
In contrast, the BLIP captioning model—designed for open-vocabulary generation—shows stronger alignment with human-provided labels. As shown in Figure~\ref{fig:caption_label_match}, upscaling improves semantic similarity across all label types: \textbf{group labels}, \textbf{supercategories}, and \textbf{affordance labels}. SwinIR and Real-ESRGAN achieve the best alignment, with Real-ESRGAN slightly leading in affordance matching and SwinIR in supercategories.

\paragraph{Match Types and Upscaling Effects.}
We evaluate three levels of alignment: \textbf{Direct Matches}: Exact string overlap (e.g., “barrel” in both caption and label). \textbf{Synonym Matches}: Lexical overlap via WordNet-style synonym sets. \textbf{Semantic Similarity}: Cosine similarity using sentence embeddings (e.g., “wizard” and “mage”).
While direct and synonym matches remain relatively sparse, semantic similarity matches increase notably after upscaling, indicating that image quality helps generate captions that conceptually align with intended labels, even when phrased differently.

\paragraph{Affordance Classification via CLIP and Multi-Label Prediction}
We frame affordance classification as a multi-label task, allowing tiles to have overlapping roles (e.g., \texttt{Characters} and \texttt{Items and Collectibles}). We use a pre-trained CLIP model (\texttt{ViT-B/32}) to extract 512-dimensional embeddings, followed by a lightweight multi-layer perceptron (MLP) with a sigmoid output layer to predict affordance types. The model is trained with binary cross-entropy loss, and a fixed threshold (e.g., 0.5) is applied at inference to assign multiple labels per tile.

\paragraph{Predicted Results.}
Table~\ref{tab:affordance_metrics} shows per-label precision, recall, and F1-score on a held-out test set. The model performs well overall, with strong results for \texttt{Characters} and \texttt{Items and Collectibles}, likely due to their visual distinctiveness and higher representation. \texttt{Interactive Object} performs lower, suggesting greater visual ambiguity. Micro- and macro-averaged F1 scores are 0.92 and 0.86, indicating robust performance across both common and rare labels.

\begin{table}[h!]
\centering
\small

\begin{tabular}{p{3cm}cccc}
\toprule
\textbf{Label} & \textbf{Prec.} & \textbf{Rec.} & \textbf{F1} & \textbf{Supp.} \\
\midrule
Characters & 0.99 & 0.96 & 0.98 & 200 \\
Environmental Obj. & 0.89 & 0.88 & 0.88 & 90 \\
Interactive Obj. & 0.50 & 0.65 & 0.56 & 31 \\
Items \& Collectibles & 0.93 & 0.97 & 0.95 & 160 \\
Terrain & 1.00 & 0.86 & 0.92 & 7 \\
\bottomrule
\end{tabular}
\caption{Per-label classification report for affordance prediction.}
\label{tab:affordance_metrics}
\end{table}


\begin{table*}[ht!]

\centering
\begin{tabular}{|c|c|c|}
\hline
\textbf{Time Frame 1} & \textbf{Time Frame 2} & \textbf{Time Frame 3} \\ \hline
Elara discovers the ancient map
&
Elara faces the treacherous paths.
&
Elara meets the Guardian dragon.
\\ 
\hline
\makecell{
    \textcolor{green}{Hollow oak} contains \\
    \textcolor{red}{ancient map} \\
    \textcolor{brown}{Elara} stands near\\
    \textcolor{green}{hollow oak}\\
    \textcolor{yellow}{Sunlight} filters through\\
    \textcolor{green}{forest canopy}
    }
&
\makecell{
    \textcolor{brown}{Elara} walks along \\
    \textcolor{gray}{rocky path} \\
    \textcolor{Orange}{Wild creatures} hide behind\\
    \textcolor{green}{dense bushes}\\
    \textcolor{yellow}{Treacherous paths} lead to\\
    \textcolor{gray}{Crystal Cavern}
    }
& 
\makecell{
    \textcolor{gray}{Crystal Cavern entrance} glows with \\
    \textcolor{yellow}{shimmering light} \\
    \textcolor{orange}{Guardian dragon} sits atop\\
    \textcolor{magenta}{crystal throne}\\
    \textcolor{brown}{Elara} stands before\\
    \textcolor{orange}{Guardian dragon}
    }
\\

\hline
\begin{minipage}[c][4cm][c]{0.3\linewidth}
\centering
\includegraphics[width=\linewidth]{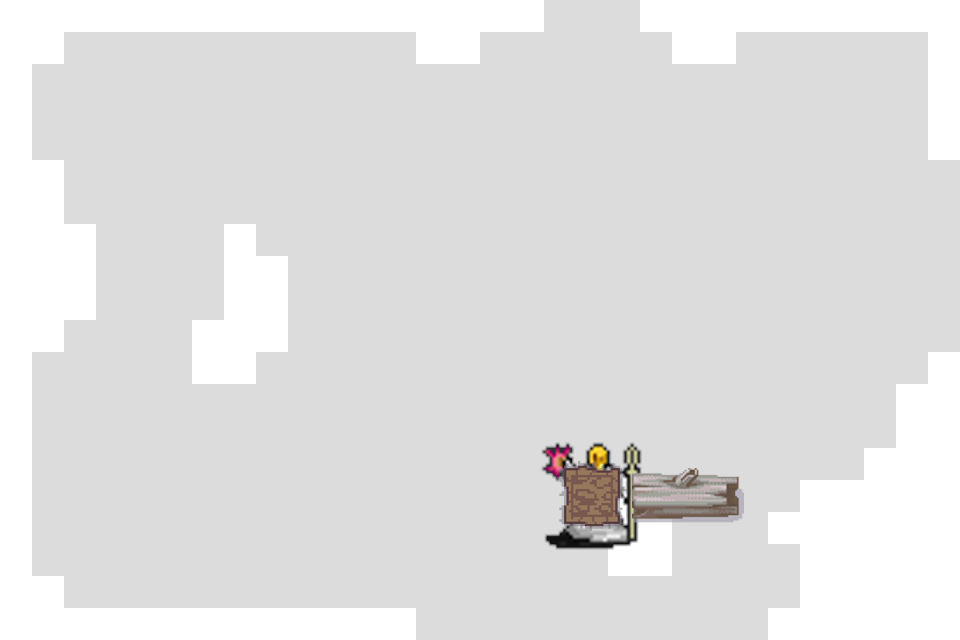}
\end{minipage}
&
\begin{minipage}[c][4cm][c]{0.3\linewidth}
\centering
\includegraphics[width=\linewidth]{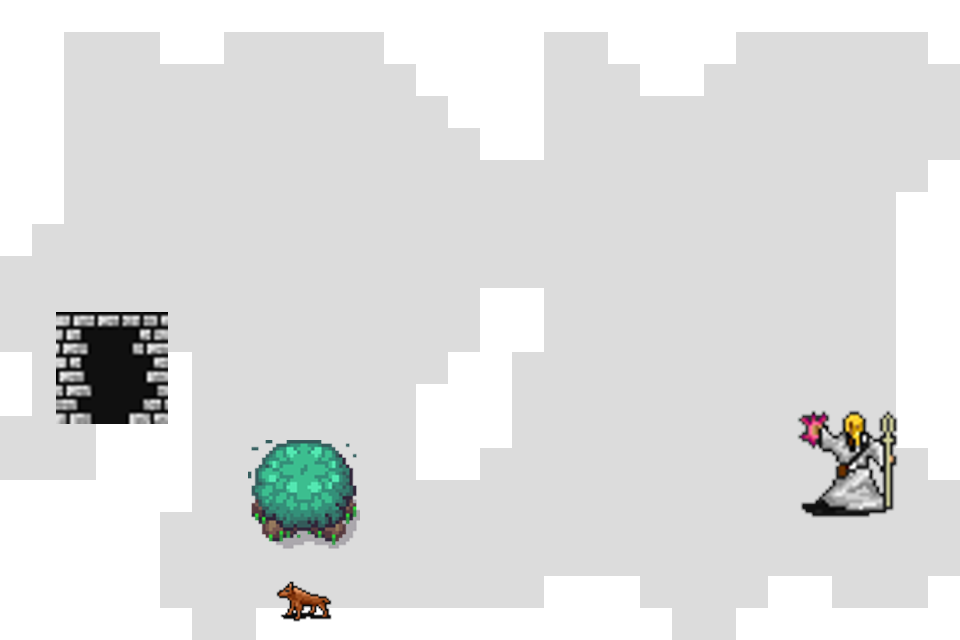}
\end{minipage}
&
\begin{minipage}[c][4cm][c]{0.3\linewidth}
\centering
\includegraphics[width=\linewidth]{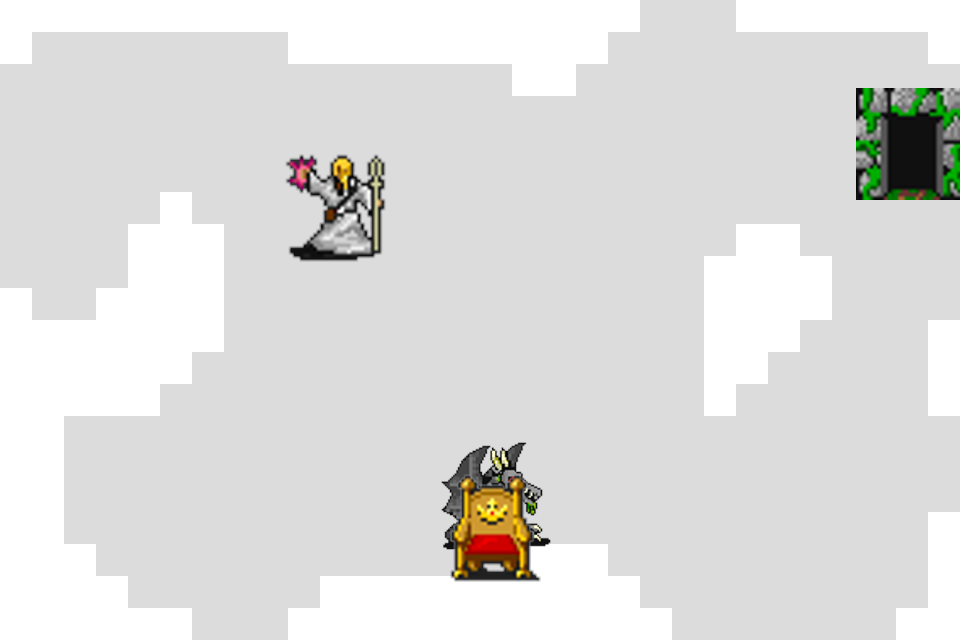}
\end{minipage}
\\
\hline

\end{tabular}
\caption{Three key time frames extracted from a generated story, represented by spatial predicates and visual renderings.}
\label{table:time_frames}
\end{table*}

%% file: 6_Experiments_Results.tex



As a proof of concept, we demonstrate how GameTileNet supports narrative-to-scene generation by combining predicate extraction, semantic tile matching, and affordance-based rendering.

{\bf Overview and Pipeline: }
We propose a pipeline that transforms narrative descriptions into semantically coherent game scenes. 
The process begins with terrain generation using Cellular Automata (CA), followed by semantic object matching, spatial placement guided by knowledge graphs (KGs), and final visual rendering.

\paragraph{Terrain Generation with Cellular Automata}
We use Cellular Automata (CA) to generate base terrain layers that form connected and navigable regions. These maps serve as the structural foundation for placing story-relevant content. CA ensures that terrain tiles are walkable and supports dynamic expansion for patch-based refinements.

\paragraph{Semantic Object Matching:}
To map narrative elements to visual assets, we construct a semantic embedding index of all annotated game tiles. Each object, annotateded with names, group labels, supercategories, and affordance types, is encoded using the \texttt{all-MiniLM-L6-v2} Sentence Transformer. During generation, narrative entities are embedded and matched to game tiles via cosine similarity, ensuring visual-semantic alignment. Affordance categories help disambiguate semantically similar objects (e.g., "guardian" as a character vs. statue).

\paragraph{Narrative Prompting and Frame Decomposition:}
Inspired by Word2World~\cite{nasir2024word2world}, we prompt an LLM to generate short adventure stories and decompose them into three key time frames using structured predicate-style descriptions. An example story and corresponding spatial breakdown are shown in Table~\ref{table:time_frames}.

\begin{itemize}
    \item Prompt\_1: "Generate a short adventure story, 100 words."
    \item Prompt\_2: "Retrieve three key time frames from the story, and describe them with [Object] [Relation] [Object] type of scene descriptions."
\end{itemize}

Relation alias, fine-tuned model with categories, Find closest tiles in the categories.

\textit{"In the heart of the Enchanted Forest, young Elara discovered an ancient map hidden within a hollow oak. It led her to the legendary Crystal Cavern, rumored to grant the finder a single wish. Braving treacherous paths and wild creatures, Elara reached the cavern's shimmering entrance. Inside, she faced the Guardian, a majestic dragon. With courage and wit, she solved the Guardian’s riddle, earning her the wish. Elara wished for peace in her war-torn village. As she exited the cavern, the skies cleared, and harmony was restored, proving that bravery and hope could transform the world."}




\paragraph{Knowledge Graph Construction:}
For each scene, we construct a knowledge graph (KG) from narrative predicates in the form \texttt{<subject> <relation> <object>}. Entities are classified by affordance types (e.g., character, item, terrain) and enriched with metadata like suggested terrain or placement confidence. Predicate relations are mapped to spatial relations (e.g., \textit{on top of}, \textit{below}) and connected through spatial and temporal edges. We build: \textbf{Scene-level KGs}, which capture objects, relations, and spatial structure. \textbf{Merged KGs}, which integrate multiple scenes and include \texttt{precedes} edges for narrative order.

\paragraph{Scene Synthesis and Rendering:}
Using the KG and terrain map, each frame is synthesized as a tile-based scene:
\begin{itemize}
    \item The terrain layer is initialized using CA with connectivity constraints.
    \item Each object is mapped to a visual tile asset and positioned using rule-based placement based on spatial relations (e.g., "above" $\rightarrow$ vertical adjacency).
    \item Object tiles are overlaid by sprite rendering, preserving size, centering, and aspect ratio.
\end{itemize}

Each scene is output as both a numerical matrix and a visual rendering. This representation enables further integration into games or interactive narrative systems. This layered narrative-to-scene pipeline bridges symbolic storytelling with 2D game synthesis. By combining semantic embeddings, spatial reasoning, and procedural terrain generation, we demonstrate a scalable approach to grounding generative narratives in visually structured environments.

While limited in scope, this module highlights the potential of GameTileNet for narrative-driven PCG. We leave a deeper evaluation of temporal consistency and agent coordination to future work in an extended study.

%% file: 8_Conclusion.tex




GameTileNet provides a structured semantic dataset tailored for low-resolution pixel art games, with annotations that support affordance prediction, narrative alignment, and symbolic scene generation. Our experiments demonstrate that combining vision-language models with upscaling techniques opens promising directions for interpreting stylized assets, particularly when conventional object detectors fall short. 

While the dataset is detailed enough for direct use in PCG pipelines, challenges remain in domain adaptation and object consistency, especially for ambiguous or multi-functional tiles. Our narrative scene generation pipeline shows how symbolic structures and spatial reasoning can bridge story elements and visual output, offering a foundation for future interactive tools or co-creative systems. We see this as a step toward grounding generative AI in playable, narrative-rich game environments. We hope GameTileNet supports researchers and developers working on PCGML, semantic asset reuse, and the integration of storytelling in 2D games.

\section*{Availability}
The GameTileNet dataset, including annotation schemas and scripts, is publicly available at \url{https://github.com/RimiChen/2024-GameTileNet}.